\documentclass{article}

\usepackage[final]{neurips_2023}

\usepackage[utf8]{inputenc} %
\usepackage[T1]{fontenc}    %
\usepackage{hyperref}       %
\usepackage{url}            %
\usepackage{booktabs}       %
\usepackage{amsfonts}       %
\usepackage{nicefrac}       %
\usepackage{microtype}      %
\usepackage{xcolor}         %
\usepackage{graphicx}
\usepackage{amsmath}
\usepackage{cleveref}       %
\usepackage{caption}        %
\usepackage{enumitem}       %
\usepackage[toc]{appendix}
\crefname{figure}{Figure}{Figures}

\title{BASE TTS: Lessons from building a billion-parameter Text-to-Speech model on 100K hours of data}

\author{
  Mateusz Łajszczak\thanks{Main contributors. Study conception and design: Mateusz, Guillermo; data collection: Mateusz, Fatih; experiments: Mateusz, Arent, Fatih, Guillermo; evaluation and analysis: Mateusz, Yang, Arent, Guillermo; draft manuscript preparation: Mateusz, Yang, Arent, Guillermo; supervision: Mateusz,Yang, Arent.} \\
\And
  Guillermo C\'ambara\footnotemark[1] \\
\And
  Yang Li\footnotemark[1] \\
\And
  Fatih Beyhan\footnotemark[1] \\
\And
  Arent van Korlaar\footnotemark[1] \\
\And
  Fan Yang \\
\And
  Arnaud Joly \\
\And
  Álvaro Martín-Cortinas\thanks{Universidad Politécnica de Madrid. Work performed while an intern at Amazon.} \\
\And
  Ammar Abbas \\
\And
  Adam Michalski \\
\And
  Alexis Moinet \\
\And
  Sri Karlapati \\
\And
  Ewa Muszyńska \\
\And
  Haohan Guo \\
\And
  Bartosz Putrycz \\
\And
  Soledad L\'opez Gambino \\
\And
  Kayeon Yoo \\
\And
  Elena Sokolova \\
\And
  Thomas Drugman \\
}

\begin{document}

\maketitle

\renewcommand{\thefootnote}{\fnsymbol{footnote}}

\begin{center}
    \Large Amazon AGI \footnote[3]{Correspondence: 
amazon-ltts-paper@amazon.com}
\end{center}
\vspace{10pt}

\renewcommand{\thefootnote}{\arabic{footnote}}
\setcounter{footnote}{0}

\begin{abstract}

We introduce a text-to-speech (TTS) model called BASE TTS, which stands for \textbf{B}ig \textbf{A}daptive \textbf{S}treamable TTS with \textbf{E}mergent abilities. BASE TTS is the largest TTS model to-date, trained on 100K hours of public domain speech data, achieving a new state-of-the-art in speech naturalness. It deploys a 1-billion-parameter autoregressive Transformer that converts raw texts into discrete codes ("speechcodes") followed by a convolution-based decoder which converts these speechcodes into waveforms in an incremental, streamable manner.
Further, our speechcodes are built using a novel speech tokenization technique that features speaker ID disentanglement and compression with byte-pair encoding.
Echoing the widely-reported "emergent abilities" of large language models when trained on increasing volume of data, we show that BASE TTS variants built with 10K+ hours and 500M+ parameters begin to demonstrate natural prosody on textually complex sentences. We design and share a specialized dataset to measure these emergent abilities for text-to-speech. We showcase state-of-the-art naturalness of BASE TTS by evaluating against baselines that include publicly available large-scale text-to-speech systems: YourTTS, Bark and TortoiseTTS. Audio samples generated by the model can be heard at \url{https://amazon-ltts-paper.com/}.

\end{abstract}

\section{Introduction}

Generative deep learning models are progressing at a rapid pace. Natural Language Processing (NLP) and Computer Vision (CV) are undergoing a fundamental shift from specialized models with supervised training, to generalized models that can achieve miscellaneous tasks with limited explicit instruction \cite{sparks}. In NLP, tasks such as question answering, sentiment analysis and text summarization can now be performed by a large language model (LLM) which was not specifically targeted for these tasks \cite{brown2020language,openai2023gpt4,scao2022bloom,touvron2023llama,hoffmann2022training}. In CV, pre-trained models that learn from hundreds of millions of image-caption pairs have achieved top performance on image-to-text benchmarks \cite{clip1,tejankar2021fistful,shen2022k}, while delivering remarkably photo-realistic results in text-to-image tasks \cite{rombach2022high, ramesh2022hierarchical, nichol2022glide, saharia2022photorealistic, yu2022scaling}. This progress has been enabled by Transformer-based architectures \cite{originaltransformerpaper} that drive improvements using many orders of magnitude more data than previous models. Similar advances are now occurring in Speech Processing and Text-to-Speech (TTS), with models leveraging thousands of hours of data that push synthesis ever closer towards human-like speech. Some of these models, described in depth in Section \ref{sec:related_work}, rely on causal language modeling tasks, like AudioLM \cite{borsos2023audiolm} or VALL-E \cite{wang2023neural}, whereas others use non-causal modules, such as SoundStorm \cite{borsos2023soundstorm} or SpeechX \cite{wang2023speechx}, or diffusion decoders \cite{NaturalSpeech2,tortoise}.

Until 2022, leading Neural TTS models were almost exclusively trained on a few hundreds of hours of recorded audio \cite{tacotron2, ecat,kim2021conditional,DBLP:conf/iclr/0006H0QZZL21,popov2021grad}. Such systems can create well-enunciated speech that is occasionally expressive for the target speakers, but typically cannot generalize beyond the small amount of training data to render ambiguous and complex texts with truly expressive spoken performance \cite{triantafyllopoulos2023overview,tahon2018can,schnell2022controllability}. To achieve such higher levels of expressiveness, TTS systems historically had to rely on labeled datasets for specific speech phenomena; and even so, achieving human-like prosody for certain types of textual inputs has remained elusive \cite{tan2021survey}. For example, in English, compound nouns and questions are notoriously hard to render correctly without accurate syntactic parsing and semantic understanding \cite{kenter2020improvingprosodywithbert}.

In this paper, we introduce BASE TTS: Big Adaptive Streamable TTS with Emergent abilities. It is a multi-lingual and multi-speaker Large TTS (LTTS) system trained on around 100K (doubling previous high in \cite{wang2023neural}) hours of public domain speech data. BASE TTS follows the approach of casting TTS as a next-token-prediction problem \cite{borsos2023audiolm,wang2023neural,tortoise}, inspired by the success of LLMs. This approach is usually applied in combination with large amount of training data to achieve strong multi-lingual and multi-speaker capabilities (e.g. one-shot voice cloning). Our goal is to improve general TTS quality and study how scaling affects the model's ability to produce appropriate prosody and expression for challenging text inputs, similar to how LLMs acquire new abilities through data and parameter scaling, a phenomenon known as "emergence" or "emergent abilities" in the LLM literature \cite{wei2022emergent, webb2023emergent}. \cite{wei2022emergent} defines \textit{emergent abilities of large language models} as "abilities that are
not present in smaller-scale models but are present in large-scale models;" for example, they show that on a range of few-shot tasks, model capability stays at a low level from 10\textsuperscript{18} to 10\textsuperscript{22} training FLOPs, but makes a drastic jump from 10\textsuperscript{22} to 10\textsuperscript{24}. To test the hypothesis that this also holds for LTTS, we propose an evaluation scheme to assess potential emergent abilities in TTS, identifying seven categories that are challenging from the literature \cite{triantafyllopoulos2023overview,tahon2018can,schnell2022controllability,tan2021survey,kenter2020improvingprosodywithbert}: compound nouns, emotions, foreign words, paralinguistics, punctuations, questions, and syntactic complexities. See more details in Section \ref{sec:evaluation}.

We design BASE TTS to model a joint distribution of text tokens followed by discrete speech representations, which we refer to as speechcodes. Discretization of speech through audio codecs \cite{vqvae,zeghidour2021soundstream,DBLP:journals/corr/abs-2210-13438} is central to our design, as it enables the direct application of methods developed for LLMs, which underlie recent works on LTTS \cite{borsos2023audiolm,wang2023neural,borsos2023soundstorm,wang2023speechx,NaturalSpeech2,tortoise, kharitonov2023speak}. Specifically, we model speechcodes using a decoder-only autoregressive Transformer with a cross-entropy training objective. Despite its simplicity, this objective can capture complex probability distributions of expressive speech, thus alleviating the oversmoothing problem seen in early neural TTS systems \cite{ren2022revisiting}. As an implicit language model, BASE TTS is also observed to make a qualitative jump in prosody rendering once a large enough variant is trained on sufficient data. 

Further, we propose speaker-disentangled speechcodes that are built on top of a WavLM \cite{wavlm} Self-Supervised Learning (SSL) speech model. We follow \cite{borsos2023audiolm} which introduces semantic tokens constructed by discretizing activations of an SSL model. We extend this approach to better control information captured by the speechcodes.
Our strategy is to limit the responsibilities of the autoregressive speechcode predictor ("speechGPT") to segmental contents, prosody, and duration, while designating a separate, speechcode-to-waveform decoder (called "speechcode decoder") with the reconstruction of speaker identity and recording conditions. We show that this convolution-based speechcode decoder is compute-efficient and reduces the whole-system synthesis time by over 70\% compared to the baseline diffusion-based decoder.

Our main contributions are summarized as follows:
\begin{enumerate}[label=\Roman*.]
  \item We introduce BASE TTS, which to our knowledge is the largest TTS model to date, featuring 1B parameters and trained on a dataset consisting of 100K hours of public domain speech data. In subjective evaluations, BASE TTS performs better than publicly available LTTS baseline models. 
  \item We demonstrate how scaling BASE TTS to larger dataset and model sizes improves its capability to render appropriate prosody for complex texts. To this end, we develop and make available an "emergent abilities" testset that can serve as a subjective evaluation benchmark for text understanding and rendering of large-scale TTS models. We report performance on different variants of BASE TTS over this benchmark, showing monotonic improvement in quality with increased dataset size and parameter count.
  \item We introduce novel discrete speech representations that are built on top of a WavLM SSL model, intended to capture only phonemic and prosodic information of the speech signal. We demonstrate that these representations outperform the baseline quantization method. We also show that they can be decoded to high quality waveforms with a simple, fast, and streamable decoder, despite high level of compression (only 400 bits/s).
\end{enumerate}

\section{BASE TTS}

\subsection{Overview}

\begin{figure}[h!]
    \centering
    \includegraphics[width=\textwidth]{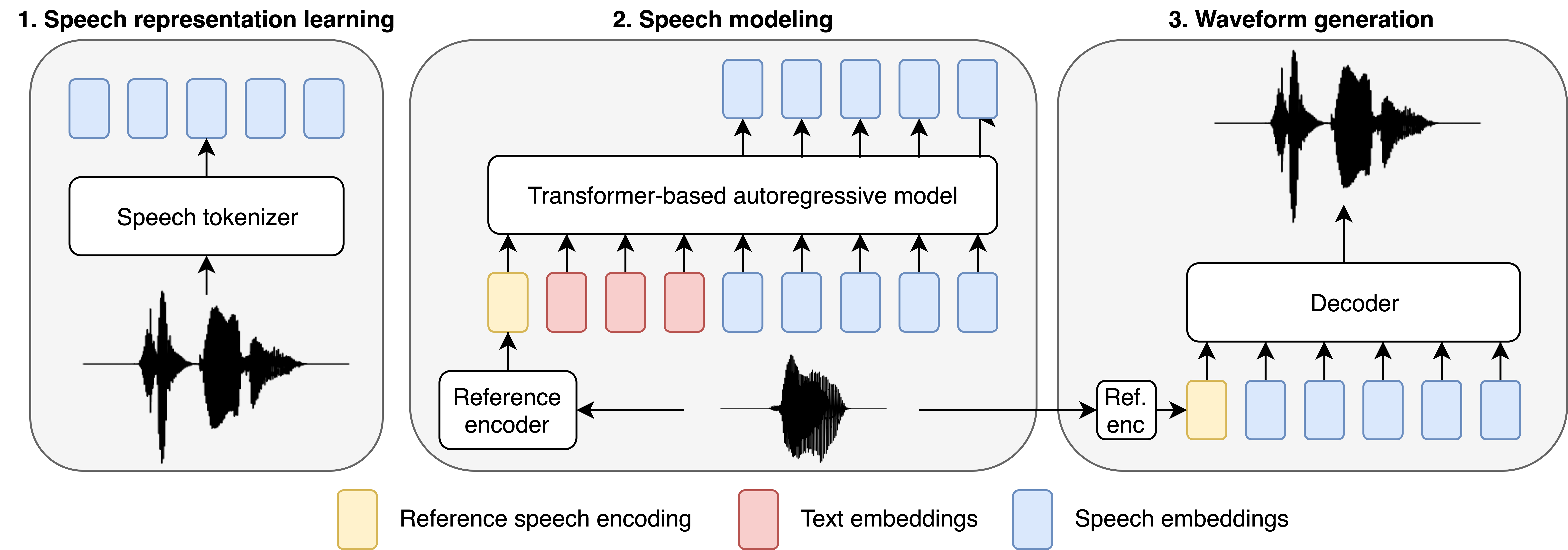}
    \caption{An overview of \textbf{BASE TTS}. The speech tokenizer (1) learns a discrete representation, which is modeled by an autoregressive model (2) conditioned on text and reference speech. The speechcode decoder (3) converts predicted speech representations into a waveform.}
    \label{fig:overview}
\end{figure}

Similar to recent works in speech modeling, we adopt an LLM-based method for the TTS task (\cref{fig:overview}).
We consider a dataset $\mathcal{D} = \{ \mathbf{x}_i, \mathbf{y}_i \}_{i=0}^N$, where $\mathbf{y}$ is an audio sample and $\mathbf{x} = \{x_1, \cdots, x_T \}$ is the corresponding text transcription. 
The audio $\mathbf{y} = \{y_1, \cdots, y_S \}$ is represented by a sequence of $S$ discrete tokens (speechcodes), learnt using a separately trained speech tokenizer. 
We use a Transformer-based autoregressive model with parameters $\phi$ in order to learn the joint probability of the text and audio sequences:

\begin{equation}
    p(\mathbf{y}, \mathbf{x}) = p(\mathbf{y} | \mathbf{x}) p(\mathbf{x}) = \prod_{s=1}^S p(\mathbf{y}_s | \mathbf{y}_{<s}, \mathbf{x}; \phi) \prod_{t=1}^T p(\mathbf{x}_t | \mathbf{x}_{<t}; \phi).
\end{equation}

The predicted speech tokens are concatenated with speaker embeddings and decoded into waveforms using a separately trained speechcode decoder consisting of linear and convolutional layers. 

\subsection{Discrete speech representations}

Discrete representation is foundational to the success of LLM, but identifying a compact and informative representation is less obvious in speech than in text, and less explored. For BASE TTS, we first experiment with a  Vector Quantized Variational Autoencoder (VQ-VAE) baseline \cite{vqvae}, where an auto-encoder based architecture reconstructs mel-spectrograms through a discrete bottleneck, as described in Section \ref{sec:vqvae}. VQ-VAE has been a successful paradigm in speech and image representation, and especially as a unit of modeling for TTS \cite{tjandra2020vqvae, tortoise}.

In Section \ref{sec:wavlm}, however, we introduce a novel method of learning speech representations through WavLM-based speechcodes. In this method, we discretize features extracted from a WavLM SSL model to reconstruct a mel-spectrogram. We apply additional loss functions to promote speaker disentanglement, and compress the resulting speechcodes using Byte-Pair Encoding (BPE) \cite{bpe} to reduce sequence length, enabling us to model longer audio with Transformers.

Both representations are compressed (325 bits/s and 400 bits/s respectively) to allow more efficient autoregressive modeling compared to popular audio codecs (e.g. 6k bits/s in \cite{wang2023neural}). With this level of compression, we aim to remove information from speechcodes that can be reconstructed during decoding (speaker, audio noise, etc.) to ensure that the capacity in speechcodes is primarily dedicated to encoding phonetic and prosodic information.

\subsubsection{Autoencoder-based speech tokens}
\label{sec:vqvae}

Our baseline discretization method is a VQ-VAE  trained to reconstruct mel-spectrograms. 
The encoder and decoder are convolutional neural networks with residual connections, which downsample the speech representation to a frequency of 25Hz. 
To (partially) disentangle speaker information from the speech representations, we introduce a global reference encoder \cite{skerry2018towards}. 
This encoder learns a fixed-size utterance-level representation, which is concatenated to the speechcodes before reconstructing with the VQ-VAE decoder.

From informal listening, we find that speechcodes produced by the autoencoder-based speech tokenizer still contain speaker information. This motivates us to develop representations with improved speaker disentanglement.

\subsubsection{WavLM-based speechcodes}
\label{sec:wavlm}
\begin{figure}[h!]
    \centering
    \includegraphics[width=\textwidth]{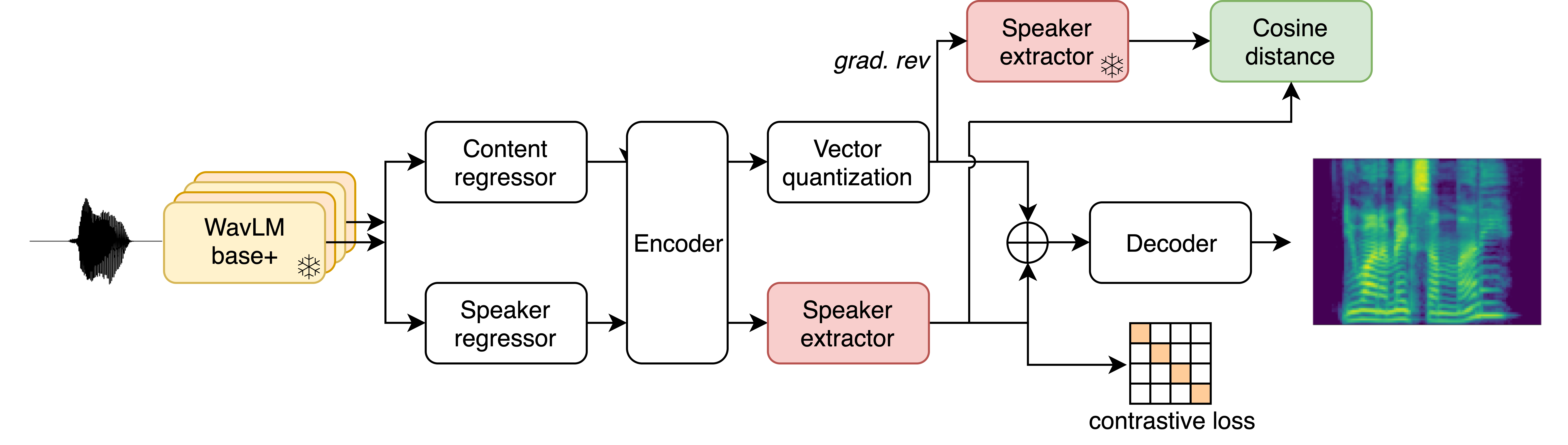}
    \caption{WavLM-based speech tokenizer. The proposed architecture encourages disentanglement of speaker and content information.}
    \label{fig:wavlm-based}
\end{figure}

We aim to develop speechcodes that contain phonetic and prosodic information, but which are disentangled from speaker identity, recording conditions, and other spurious features in the audio signal. To this end, we introduce a speech tokenizer based on features extracted from a pretrained WavLM model \cite{wavlm}, further trained with losses that encourage disentangling the speaker identity. Our approach similar to the one introduced in \cite{martíncortinas2024enhancing} with modifications that reduce bitrate of the codes.
The overall architecture of the speech tokenizer is shown in \cref{fig:wavlm-based}.

We first pass the waveform through the WavLM model and extract the hidden states. 
These hidden states are then passed through separate content and speaker linear regressors. 
The output of these regressors is then fed into a convolutional residual encoder \cite{he2016deep}. 
The content encodings are passed through a vector quantization module that outputs one speechcode per one WavLM frame (i.e. 20ms of speech). 

The speaker encodings are passed through a Transformer-based speaker extractor \cite{originaltransformerpaper} to obtain the speaker embeddings. The model only extracts, and we only use non-specific features that cannot be used for identification.

The speaker embeddings are concatenated with the speechcodes, and decoded into a spectrogram using a convolutional decoder. We then compute L1 distance between decoded and target spectrograms and use it as the reconstruction loss. While L1 is not the optimal reconstruction objective, we prioritize representations that are conducive for autoregressive modeling \cite{de2019hierarchical}, and demonstrate accordingly that the final audio quality can be kept high when this learned representation is decoded with our speechcode decoder, in Section \ref{sec:decoder}.
The speaker embeddings are used in a contrastive loss, maximizing the similarity between samples from the same speaker and minimizing it for those from different speakers \cite{oord2018representation}.
Furthermore, we maximize the cosine distance between the speaker embeddings and embeddings obtained by passing the output of the content regressor through the frozen speaker extractor and applying gradient reversal \cite{ganin2015unsupervised}. 
We hypothesize that this encourages disentanglement between content and speaker information. 

In addition to better disentanglement of speaker information, we also believe that using features from a pretrained WavLM model as input (as opposed to a jointly learnt audio encoding) keeps speechcodes more robust to recording conditions. Our intuition is that WavLM was trained with data augmentation to encourage disentanglement from background noise.
The total loss is given by a weighted combination of these losses, in addition to the commitment loss for the vector quantizer: 

\begin{equation}
\label{eq:wavlm}
    L = L_\text{recon} + \alpha L_\text{commitment} + \beta L_\text{contrastive} + \gamma L_\text{cosine}
\end{equation}

\subsubsection{Byte-pair encoding on speechcodes}

The WavLM-based speech representations are learned at a frequency of 50 Hz to ensure that they offer enough resolution in time to be able to discriminate between single phonemes even in fast-paced speech (the average length of English phonemes tends not to dip below 20ms \cite{kuwabara1996acoustic}).
We apply Byte-Pair Encoding \cite{bpe} to reduce the average sequence length of speechcodes by around 40\%, in order to mitigate the quadratic memory increases in Transformers sequence length and minimize complexity for the autoregressive model. Similar to the BPE of texts, we iteratively join the most frequent pairs of speechcodes into new tokens and add them to the vocabulary until a pre-determined vocabulary size is reached.

\subsection{Autoregressive speech modeling (SpeechGPT)}
\label{sec:speechGPT}

We train a GPT2-architecture autoregressive model \cite{radford2019language} that we call "SpeechGPT" to predict the speechcodes conditioned on text and reference speech. 
The reference speech conditioning consists of a randomly selected utterance from the same speaker, which is encoded to a fixed-size embedding. 
The reference speech embedding, text, and speechcodes are concatenated into a single sequence that is modeled by a Transformer-based autoregressive model. 
We use separate positional embeddings and separate prediction heads for text and speech.

We train the autoregressive model from scratch, without pretraining on text (e.g. as done in \cite{twist}). In order to retain textual information to guide prosody, we also train SpeechGPT with an objective to predict the next token in the text portion of the input sequence, so that speechGPT is partially a text-only LM. We apply a lower weight to this text loss compared to the speech loss.

\subsection{Waveform generation}
\label{sec:decoder}

Our baseline uses a diffusion-based spectrogram decoder and a separately trained UnivNet \cite{univnet} vocoder, as proposed in \cite{tortoise}. Diffusion-based TTS decoding can generate high-quality speech, but it suffers from slow inference and cannot generate samples incrementally - This lack of streamabilility forces us to obtain the audio output for the entire sequence in one go.  
Furthermore, the diffusion model in \cite{tortoise} predicts spectrograms, and requires a separately trained vocoder to generate audio, complicating the training and inference pipeline. 

\begin{figure}[h!]
    \centering
    \includegraphics[width=0.8\textwidth]{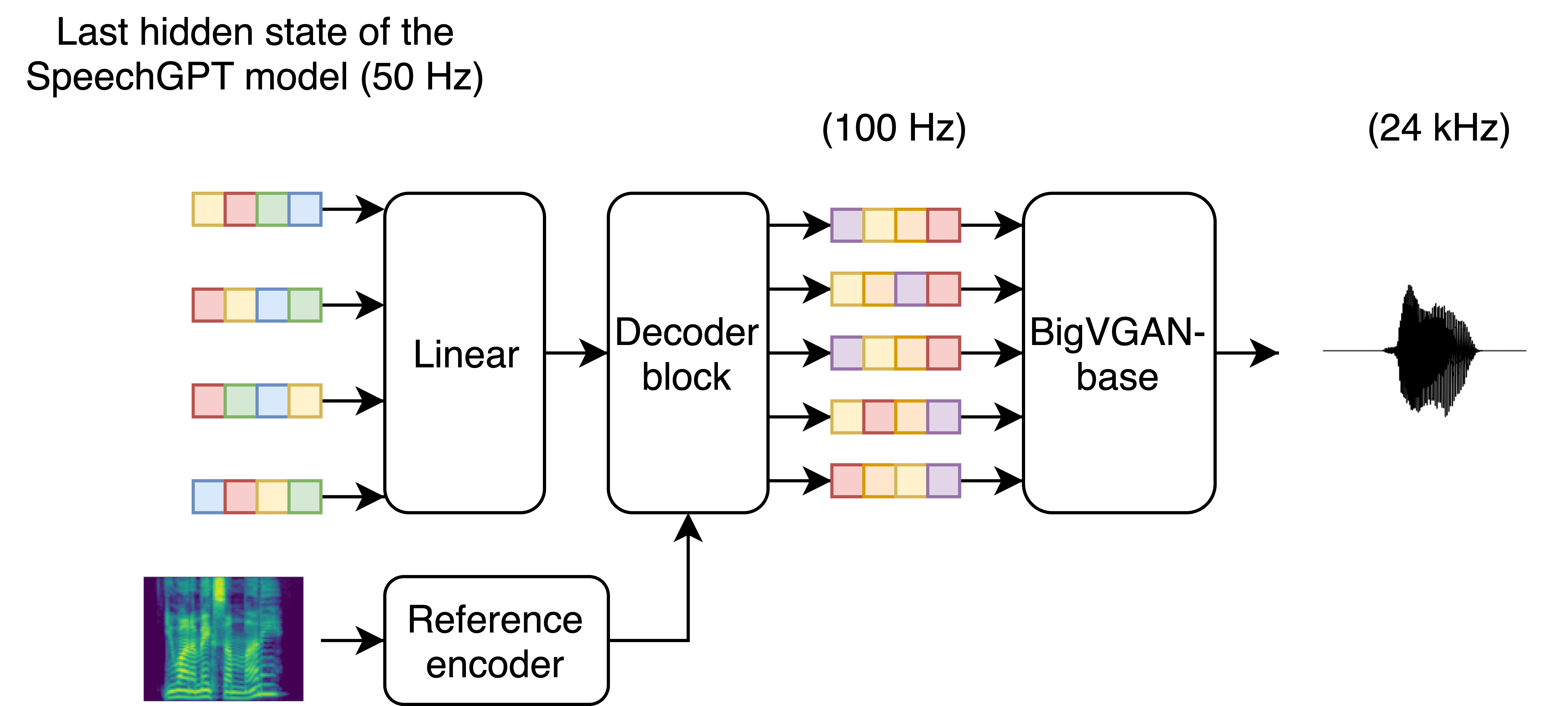}
    \caption{Speechcode decoder. The last intermediate representations of the SpeechGPT model are upsampled by a factor 2 in a decoder block. BigVGAN-base vocodes these representations into the waveform.}
    \label{fig:scv}
\end{figure}

Our proposed decoder, inspired by \cite{ecat}, is trained in an end-to-end fashion to predict waveforms. Our variant uses speechcodes as an input to the model instead of phoneme encodings and prosody latents. Additionally, to make the model more scalable, we replace LSTM layers \cite{hochreiter1997long} with convolutional ones to decode an intermediate representation. 
The output of a HiFi-GAN based decoder block \cite{hifigan} is fed into BigVGAN vocoder \cite{bigvgan} to predict the waveform. In training, we use the same set of adversarial and non-adversarial losses as \cite{ecat}. We call our proposed system a "speechcode decoder" and depict it in \cref{fig:scv}. In addition to simplifying the overall system, we hypothesize that training the decoder and vocoder end-to-end yields higher-quality speech.

In practice, instead of speechcodes, the speechcode decoder takes as an input the last hidden state of the autoregressive Transformer. 
We do so because the dense latent representation provides much richer information than a single speechcode, following \cite{tortoise}.
During training, we feed the text and target codes to the trained SpeechGPT (with parameters frozen) and then condition the decoder on the last hidden state.
Feeding the last hidden states of SpeechGPT helps improving the segmental and acoustic quality of speech, but also couples the decoder to a specific version of SpeechGPT.
This complicates experimentation because it forces the two components to  be always built sequentially. 
This limitation needs to be addressed in future work.

\section{Experimental setup}
We design and conduct experiments to validate the architecture of BASE TTS and its quality, capabilities and compute performance. First, we compare model quality achieved by autoencoder-based and WavLM-based speechcodes. We then evaluate the two methods of acoustically decoding the speechcodes: a diffusion-based decoder and a speechcode decoder. Having completed these architectural ablations, we assess the emergent abilities of BASE TTS in 3 variants of dataset sizes and model parameters, assessed by an expert linguist. Further, we run subjective MUSHRA tests to measure naturalness, as well as automated intelligibility and speaker similarity measurements. We also report the quality of speech comparison against other open-source text-to-speech models. 

\subsection{Dataset}\label{dataset}

In order to test our hypothesis that abilities emerge with the scale of data, we set out to train with the largest speech dataset for our largest LTTS model. We create a dataset consisting of 100K hours of unlabeled, public domain speech data. The overall dataset is dominated by English data (over 90\%), followed by German, Dutch, Spanish. We first download mp3 files from the Web, and resample them into 24 kHz mono LPCM files, with 16 bits per sample encoded as signed-integers. The vast majority of the dataset is recorded in non-studio conditions and contain noises, and we avoid additional signal processing or denoising in order to test our model's ability to generate clean speech from noisy background data.

Next, we use an ASR model combined with Voice Activity Detection (VAD) to perform speech segmentation and Automatic Speech Recognition. The ASR model splits the entire speech into 30-second or shorter fragments. We then perform VAD to split sentences longer than 20 seconds, and reassemble them into longer sentences during training as long as their total length does not exceed 40 seconds. We thus expose BASE TTS to speech segments which are between 0 and 40 seconds long, containing one or more sentences, so that the model becomes robust to both very short inputs, and learns from longer context \cite{9054106}. 

To generate the transcript for the dataset, we initially relied on the ASR transcription. We found that ASR tends to generate a smaller set of punctuations than their natural occurrences. For example, it over-generates commas instead of colons and semicolons, and rarely generates parenthesis. We then perform partial text restoration (similar to \cite{kang2023libriheavy}) by: 1) searching for source texts from the Internet and tying them to each recording; 2) matching the ASR transcript with the source text sentence by sentence, replacing the former with latter if their difference is small enough (length no longer/shorter by a factor of 3 and within a certain edit distance). We restore around 1/2 of total texts to the `original', resulting in an 300\% increase in quotation marks of all kinds and 20000\% increase in parentheses in the transcript. This led to our models no longer having frequent acoustic artefacts on parentheses and other `rare' punctuations. 

\subsection{Training and hyperparameters}

We train the models in three steps. Due to the use of discrete speech representations, we rely on the cross-entropy validation loss on $\sim$50 hours held-out from training data to guide our training, restarting, and stopping decisions. All models are trained on clusters of AWS P3dn instances with NVIDIA® V100 GPUs. 

First, we train two speech tokenizer variants: a VQ-VAE tokenizer as in Section \ref{sec:vqvae} and a WavLM tokenizer as in Section \ref{sec:wavlm}. Due to the overwhelming presence of English data, we train both variants on a subset of the data in Section \ref{dataset} by keeping all data from non-English languages, but capping the amount of speech from individual English speakers at 200 hours - excess data from the speaker are randomly discarded. We use a codebook size of 256 for the WavLM speechcodes and 8196 for the VQ-VAE ones. The WavLM speechcodes are subsequently compressed with the BPE algorithm and a vocabulary of size 8192. We first optimize the reconstruction loss on ground truth spectrograms, sanity-checking the reconstructed audio on "minority" languages such as Polish (0.2\% of total data) to ensure sufficient speechcode coverage. 

Second, we train SpeechGPT on the entire training dataset with random initial weights. The inputs to the model are: random reference speech segment from a speaker, text and corresponding target speech segments. We convert the reference speech segment to a mel-spectrogram and feed it to the reference encoder. The target speech segment is tokenized by either VQ-VAE or WavLM, forming the target speechcodes. We concatenate the encoded reference with input text embeddings and speechcodes embedding, and feed them to the autoregressive Transformer. The model is optimized with cross-entropy loss on both text tokens (weight 0.01) and speechcodes (weight 1.0).  

Our autoregressive Transformer structure is largely identical to the GPT-2 language model \cite{radford2019language}. For most of the experiments reported below ("BASE-large"), we train a 32-layer decoder-only transformer with masked self-attention heads (1536-dimension states and 24 attention heads). For the feed-forward networks, we use 6144-dimension inner states. We use Adam \cite{kingma2017adam} optimization with a max learning rate of 3.0e-4 and weight decay of 0.03. The learning rate was increased linearly from zero over the first 10000 updates and annealed to 1.5e-4 using a cosine schedule.

To understand the importance of data and model size, we build two other versions (BASE-small \& BASE-medium) of SpeechGPT with WavLM speechcodes as described in Section \ref{sec:wavlm}, with increasing number of parameters and speech data in Table \ref{table:ablation-exp-table}.

\begin{table}[htp]
    \centering
    \caption{Ablation experiments on speechGPT with 3 data sizes and parameters}
    \vspace{1em}
    \label{table:ablation-exp-table}
    \begin{tabular}{llll} %
        \toprule
        \textbf{Attributes} & \textbf{BASE-small} & \textbf{BASE-medium} & \textbf{BASE-large} \\
        \midrule
        \textbf{Data amount} & 1K hours & 10K hours & 100K hours \\
        \textbf{Parameters in SpeechGPT} & 150 million  & 400 million & 980 million \\
        \textbf{Attention heads} & 12 & 16 & 24 \\
        \textbf{Transformer layers \& dims} & 16 layers, 768 dims & 30 layers, 1024 dims & 32 layers, 1536 dims \\
        \textbf{Feed-forward dims} & 3072 & 4096 & 6144 \\
        \textbf{Speechcode} & wavLM & wavLM & wavLM \\
        \textbf{Decoder} & Speechcode decoder & Speechcode decoder & Speechcode decoder \\
        \bottomrule
    \end{tabular}
    \vspace{1em}
\end{table}

Finally, we train the speechcodes decoder described in Section \ref{sec:decoder} with a modified log-likelihood-maximization GAN training scheme borrowed from \cite{ecat}. Total parameter size for this module is around 150 million.

\subsection{Evaluation methodology}
\label{sec:evaluation}

Three types of tests are considered to assess the model quality:

\begin{enumerate}[label=\Roman*.]
  \item \textbf{MUSHRA (Multiple Stimuli with Hidden Reference and Anchor)} among competing systems by native listeners in US English and Spanish. We use a MUSHRA scale ranging from 0 to 100. We do not place an Upper/Lower anchor or ask listeners to rate those at 0 or 100. This way, we are able to get more ratings on target systems on a fixed evaluation budget. For each MUSHRA test reported, 25-50 testers provide ratings on 50-100 text snippets that range between 5-40 seconds long, such that we achieve about 17-20 hours of total effective listening time per system per test. 
  \item \textbf{Linguistic expert evaluation of "emergent abilities".} To gauge the ability of BASE TTS to achieve finer understanding of the text, we hand-created an "emergent abilities testset" in English with 7 categories of texts: Questions, Emotions, Compound Nouns, Syntactic Complexities, Foreign Words, Paralinguistics, and Punctuations. In Table \ref{table:emergent-abilities}, we present an example from each category, and how a linguistic expert rates the TTS output on a discrete 3-point scale. These sentences are designed to contain challenging tasks - parsing garden-path sentences \cite{garden-path}, placing phrasal stress on long-winded compound nouns \cite{syntax-prosody}, producing emotional or whispered speech, or producing the correct phonemes for foreign words like "qi" or punctuations like "@" - none of which BASE TTS is explicitly trained to perform. Our hypothesis is that as BASE TTS increases in model capacity and trains over more data, the model will start to acquire these abilities, following evidence that scaling in these dimensions begets qualitative ability jumps \cite{sparks, webb2023emergent, wei2022emergent}. We share the full testset in Appendix \ref{appendix:emergent}.
  \item \textbf{Automated objective evaluations} to test TTS robustness, especially issues with missed, duplicated or hallucinated synthesis \cite{NaturalSpeech2}, and speaker similarity.  We use an ASR model to compute the Word Error Rate (WER) by comparing the testset text (ground truth) against the ASR output from the synthetic speech. 
In addition, we employ a speaker verification model
\footnote{\url{https://huggingface.co/microsoft/wavlm-base-plus-sv}} fine-tuned on WavLM features \cite{wavlm} to obtain speaker embeddings from the original recordings and synthetic speech, and then we compute the cosine distance between them to get a speaker similarity metric (SIM). 
\end{enumerate}

\begin{table}[htp]
    \centering
    \caption{Emergent abilities testset by category and evaluation criteria.}
    \vspace{1em}
    \label{table:emergent-abilities} 
    \begin{tabular}{p{1.8cm}p{5cm}p{5.5cm}} %
        \toprule
        \textbf{Categories} & \textbf{Example sentence} & \textbf{Evaluation criteria} \\
        \midrule
        \textbf{Compound Nouns} & The Beckhams decided to rent a charming stone-built quaint countryside holiday cottage. & 1 = fails to recognise compound nouns \newline 2 = fails to realise the phrasal stress naturally \newline 3 = natural phrasal stress \\ [2ex]
        \textbf{Emotions} & "Oh my gosh! Are we really going to the Maldives? That’s unbelievable!" Jennie squealed, bouncing on her toes with uncontained glee.  &   1 = no audible emotions \newline 2 = emotion present but insufficient \newline 3 = correct emotion recognition and appropriate rendering \\ [2ex]
        \textbf{Foreign Words} & Mr. Henry, renowned for his mise en place, orchestrated a seven-course meal, each dish a pièce de résistance. & 1 = pronounces foreign words with incorrect anglicized pronunciation \newline 2 = applies foreign accent but not entirely correctly \newline 3 = correct rendering in the intended language or accepted anglicized reading  \\ [2ex]
        \textbf{Paralinguistics} & "Shh, Lucy, shhh, we mustn't wake your baby brother," Tom whispered, as they tiptoed past the nursery. & 1 = no recognition of paralinguistic keywords such as "shhh" or "phew" \newline 2 = clear intention to render keywords distinctly, but rendering unnatural \newline 3 = natural rendering, e.g. making speech voiceless on "shhh" and other whispered speech   \\ [2ex]
        \textbf{Punctuations} & She received an odd text from her brother: 'Emergency @ home; call ASAP! Mom \& Dad are worried...\#familymatters.' & 1 = glitches on uncommon punctuations such as \# or \& \newline 2 = no glitch but incorrect rendering \newline 3 = no glitch and correct pausing and verbalization, e.g. @ as "at".  \\ [2ex]
        \textbf{Questions} & But the Brexit question remains: After all the trials and tribulations, will the ministers find the answers in time? & 1 = intonation pattern incorrect \newline 2 = intonation pattern largely correct but with minor flaws \newline 3 = correct intonation  \\ [2ex]
        \textbf{Syntactic Complexities} & The movie that De Moya who was recently awarded the lifetime achievement award starred in 2022 was a box-office hit, despite the mixed reviews. & 1 = failure to parse the syntax correctly \newline 2 = parses the syntax largely correctly but the rendering is not entirely natural \newline 3 = parsing correct and rendering natural \\
        \bottomrule
    \end{tabular}
    \vspace{1em}
\end{table}

For subjective evaluations, we report average MUSHRA scores with $95\%$ confidence intervals. Furthermore, to determine the significance of differences between two systems, we conduct a t-test; if the p-value is $< 0.05$, we consider the difference significant. To aid visualization we mark statistically significantly better systems with \textbf{bold}.

\section{Results}

\subsection{VQ-VAE speechcode vs. WavLM speechcodes}

We conduct MUSHRA evaluations on 6 US English and 4 Spanish (Castilian and US) speakers in order to test comprehensively the quality and generalisability of the two speech tokenization approaches, using both held-out test data for speakers seen in the training ("seen" condition) and unseen speakers ("one-shot" condition). 
In terms of average MUSHRA scores for English voices, VQ-VAE and WavLM based system are on par (VQ-VAE: 74.8 vs WavLM: 74.7, the difference is statistically non-significant). However, for Spanish voices, WavLM based model outperforms the VQ-VAE one (VQ-VAE: 73.3 vs WavLM: 74.7) in a statistically significant way. Note that English data comprises around 90\% of our dataset, while Spanish data only 2\%. We hypothesize that speechcode improvements are more critical for low-resource languages while sheer data volume can make up for imperfect representations. Further verification of this hypothesis needs to be addressed in future work. Since WavLM-based system performs at least as well or better as the VQ-VAE baseline, we use it to represent BASE TTS in further experiments. In Table \ref{VQVAE-WAVLM-TABLE}, we show the results categorized by speaker to provide additional details on per-speaker performance.

\begin{table}[htp]
  \caption{Results of MUSHRA evaluation comparing VQ-VAE and WavLM speechcodes. We report mean scores for 10 US English \& Spanish speakers and highlight statistically significant winners with \textbf{bold}.}
  \label{VQVAE-WAVLM-TABLE}
  \centering
  \begin{tabular}{lcccc}
    \cmidrule(r){1-5}
    \textbf{Speaker} & \textbf{Language} & \textbf{One-shot or seen} & \textbf{VQ-VAE} & \textbf{WavLM} \\
    \midrule
    Male speaker A & US English & seen & $76.3~\pm~0.6$ & $76.3~\pm~0.6$ \\ %
    Male speaker B & US English & one-shot & $76.3~\pm~0.6$ & $76.3~\pm~0.6$ \\ %
    Male speaker C & US English & one-shot & $73.6~\pm~0.6$ & $73.4~\pm~0.6$  \\ %
    Female speaker A & US English & seen & $74.5~\pm~0.7$ & $74.5~\pm~0.7$  \\ %
    Female speaker B & US English & seen & $73.8~\pm~0.7$ & $74.0~\pm~0.7$  \\ %
    Female speaker C & US English & one-shot & $74.1~\pm~0.6$ & $73.9~\pm~0.6$  \\ %
    Male speaker D & Castilian Spanish & seen & $72.5~\pm~0.9$ & $\textbf{74.4}~\pm~0.9$ \\ %
    Male speaker E & US Spanish & one-shot & $73.9~\pm~0.8$ & $74.4~\pm~0.8$  \\ %
    Female speaker D & Castilian Spanish & seen & $77.0~\pm~0.7$ & $77.0~\pm~0.7$ \\ %
    Female speaker E & Castilian Spanish & one-shot & $68.5~\pm~1.0$ & $\textbf{72.4}~\pm~1.0$  \\ %
    \bottomrule
  \end{tabular}
\end{table}

\subsection{Diffusion-based decoder vs. speechcode decoder}

BASE TTS simplifies over the baseline diffusion-based decoder by proposing an end-to-end speechcode decoder, as described in Section \ref{sec:decoder}. Our approach offers streamability and 3X improvement in inference speed. In order to ensure that this approach does not degrade quality, we run an evaluation of the proposed speechcode decoder against the baseline. Table \ref{DECODER-TABLE} presents the results of MUSHRA evaluation we conducted for 4 US English and 2 Spanish speakers. For 4 voices out of 6, the BASE TTS variant with speechcode decoder outperforms the diffusion-based baseline in terms of average MUSHRA score. For the remaining speakers the difference is not statistically significant. We conclude that the speechcode decoder is the preferred approach, as it does not degrade quality and for most of the voices it brings quality improvements, while offering faster inference. Our results suggest that combining two powerful generative models for speech modeling is redundant and can be simplified by dropping the diffusion decoder.

\begin{table}[htp]
  \caption{Results of MUSHRA evaluation comparing diffusion-based decoder and speechcode decoder. We report mean scores for 6 US English \& Spanish speakers and highlight statistically significant winners with \textbf{bold}.}
  \label{DECODER-TABLE}
  \centering
  \begin{tabular}{lcccc}
    \cmidrule(r){1-5}
    \textbf{Speaker} & \textbf{Language} & \multicolumn{1}{p{2cm}}{\centering \textbf{One-shot \\ or seen}} &  \multicolumn{1}{p{3cm}}{\centering \textbf{Diffusion-based \\ decoder}} & \multicolumn{1}{p{2cm}}{\centering \textbf{Speechcode \\ decoder}} \\
    \midrule
    Male speaker A & US English & seen & $73.3~\pm~0.8$ & $\textbf{74.9}~\pm~0.8$  \\ %
    Male speaker B & US English & one-shot & $74.6~\pm~0.7$ & $\textbf{75.5}~\pm~0.7$ \\ %
    Female speaker A & US English & seen & $77.9~\pm~0.7$ & $77.4~\pm~0.7$  \\ %
    Female speaker C & US English & one-shot & $69.2~\pm~0.8$ & $\textbf{71.1}~\pm~0.8$  \\ %
    Male speaker D & US Spanish & seen & $77.3~\pm~0.8$ &	$77.0~\pm~0.8$ \\ %
    Female speaker E & Castilian Spanish & one-shot & $72.8~\pm~1.0$ & $\textbf{74.1}~\pm~1.0$ \\ %
    \bottomrule
  \end{tabular}
\end{table}

\subsection{Emergent abilities - Data and model size ablation}

In this section, we report on our verification of the hypothesis that data and parameter scaling in LTTS brings qualitatively different results, analogous to training LLMs from 10\textsuperscript{22} to 10\textsuperscript{24} tokens, when LLMs "suddenly" start to master few-digit addition, recognise words in context at above chance level, and transcribe speech with expert phonetic alphabet.

We perform both MUSHRA and Linguistic expert judgement of "Emergent abilities," as described in \ref{sec:evaluation}, on 2 American English voices. We report all scores by BASE-small, BASE-medium and BASE-large systems, as outlined in Table \ref{table:ablation-exp-table}. In the MUSHRA results in Table \ref{ablation-table}, we observe that speech naturalness improves significantly from BASE-small to BASE-medium, but less so from BASE-medium to BASE-large - the difference is statistically significant only in Male speaker A.

\begin{figure}[h!]
    \centering
    \includegraphics[width=\textwidth]{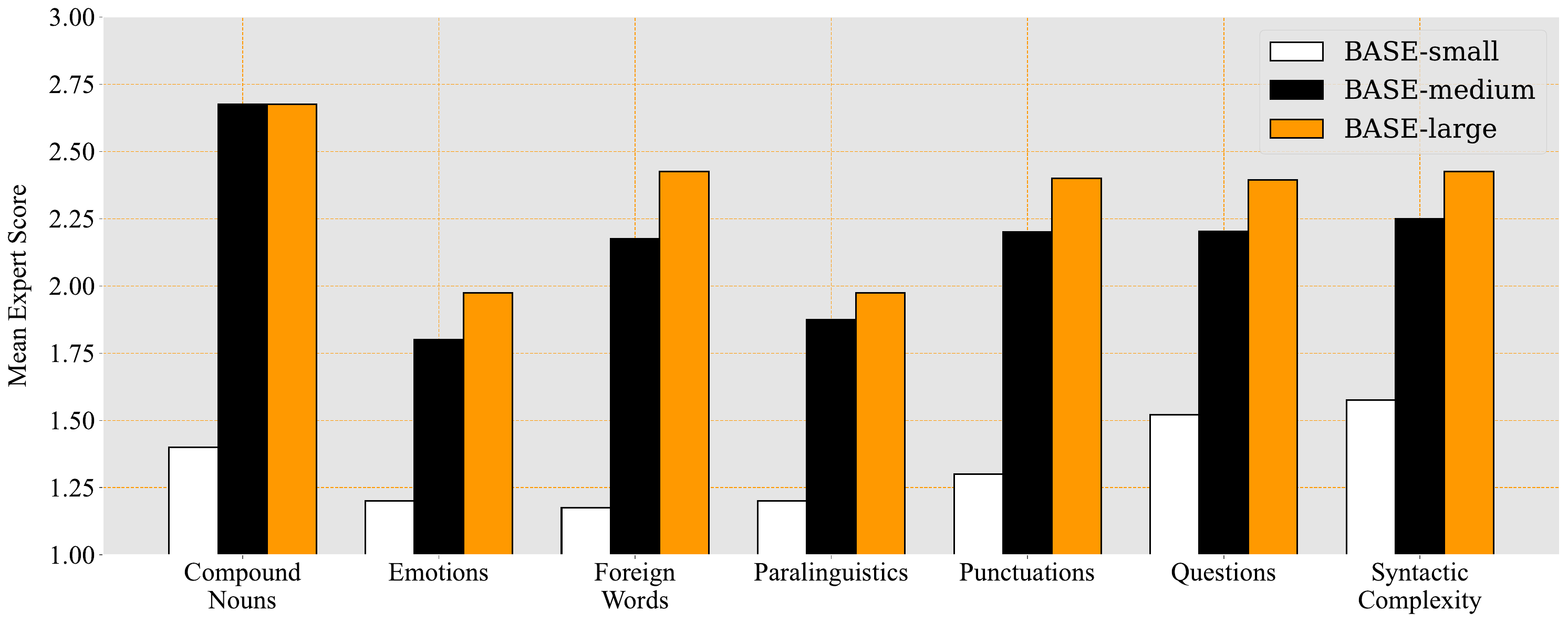}
    \caption{Linguist expert evaluations per system: BASE TTS - small/medium/large. Results are presented for the seven proposed tasks, by computing the mean of the expert scores over 20 sentences in each category.}
    \label{fig:jobInformationDialog}
\end{figure}

We report the results of the linguistic expert judgement for the three systems, with their average score for each category, in Figure \ref{fig:jobInformationDialog}. First, we see a universal jump from BASE-small to BASE-medium across categories. BASE-small appears fundamentally unable to interpret emotions, paralinguistics, and foreign words (average score < 1.25, score lower bounded at 1), and never reaches an average score of 1.75 in any category. By contrast, at BASE-medium, the model has mastered compound nouns and makes a significant jump in all categories; BASE-medium never performs below an average score of 1.75 in any category. From BASE-medium to BASE-large, we observe continued but diminishing improvement in all categories except compound nouns, where the performance has saturated. Combined with the findings from naturalness MUSHRA, we believe that scaling GPT-based TTS model from 1000+ to 10000+ hours and model size from 100 million to 500 million is the point at which "emergent abilities" \cite{wei2022emergent} start to occur for our TTS. 

Examining results category by category, we observe that Emotions and Paralinguistics remain the most challenging tasks for all model variants, and even BASE-large performs at around an average score of 2.0, which indicates that model can recognise and react to relevant keywords, but the prosody quality remains imperfect to expert judgement. By contrast, BASE-medium can perform close to ceiling on English Compound Nouns, while trailing BASE-large slightly in most of the other categories (Foreign Words, Punctuations, Questions, Syntactic Complexity). We remain hopeful that further scaling and injection of textual knowledge from text-only LLM can help us close remaining performance gaps.

\begin{table}[htp]
  \caption{Results of MUSHRA evaluation comparing 3 data-parameter settings on emergent abilities testset. We report mean scores for 2 US English speakers with statistically significant winners highlighted in \textbf{bold}.}
  \label{ablation-table}
  \centering
  \begin{tabular}{lccc}
    \cmidrule(r){1-4}
    \textbf{Speaker} & \textbf{BASE-small} & \textbf{BASE-medium} & \textbf{BASE-large} \\
    \midrule
    Male speaker A & $61.5~\pm~0.7$	& $70.5~\pm~0.6$	& $\textbf{72.2}~\pm~0.5$ \\
    Female speaker A & $68.1~\pm~0.6$ & $\textbf{72.0}~\pm~0.6$ & $\textbf{72.3}~\pm~0.6$ \\
    \bottomrule
  \end{tabular}
\end{table}

\subsection{BASE TTS vs. industry baselines}

We select three industry baselines with publicly available pre-trained models: YourTTS \cite{casanova2022yourtts}, Bark\footnote{\url{https://github.com/suno-ai/bark}}, and Tortoise \cite{tortoise}. We conduct the comparison exclusively in one-shot setting, using 10-second reference clip for two US English speakers which BASE TTS was not previously trained or evaluated on. 

As seen in Table \ref{competitor-mushra}, BASE TTS obtains significantly higher naturalness scores in MUSHRA than baseline systems, among two US English speakers tested.

We further conduct objective evaluation on WER and speaker similarity on 10 speakers and 307 samples for each, reported in Table \ref{table:objective_eval}. Overall, BASE TTS produces the most natural speech, the least amount of misalignment with input text, and the most similar speech to the reference speaker, followed relatively closely by Tortoise on naturalness and WER. Bark and YourTTS perform far worse on naturalness and WER, although Bark is relatively close to BASE TTS in speaker similarity. 

\begin{table}[htp]
  \caption{Results of MUSHRA evaluations comparing BASE TTS with industry baselines. We report mean scores for 2 US English speakers with statistically significant winners highlighted in \textbf{bold}.}
  \centering
  \label{competitor-mushra}
  \begin{tabular}{lcccc}
    \cmidrule(r){1-5}
    \textbf{Speaker} & \textbf{BASE TTS} & \textbf{Tortoise} & \textbf{Bark} & \textbf{YourTTS} \\
    \midrule
    Male speaker A & $\textbf{71.7}~\pm~1.0$ & & $46.0~\pm~1.0$ & $47.6~\pm~1.0$ \\ %
    Male speaker A & $\textbf{73.7}~\pm~0.9$ & $68.8~\pm~0.8$ &  &  \\ %
    Female speaker A & $\textbf{67.8}~\pm~1.0$ &  & $49.7~\pm~1.0$ & $39.7~\pm~1.0$ \\ %
    Female speaker A & $\textbf{68.2}~\pm~1.0$ & $58.5~\pm~1.0$ & &  \\ %
    \bottomrule
  \end{tabular}
\end{table}

\begin{table}[htp]
    \centering
    \caption{Word Error Rate (WER) and Speaker Similarity (SIM) metrics in percentages (\%) \\  for our models and industry baselines. We report means and highlight winners with \textbf{bold}.}
    \vspace{0.5em}
    \label{table:objective_eval}
    \begin{tabular}{cccccc} %
        \toprule
          & \textbf{BASE TTS} & \textbf{Bark} & \textbf{Tortoise} & \textbf{YourTTS} \\
        \midrule
        \textbf{WER} $\downarrow$ & $\textbf{6.5}$ & $19.2$ & $8.0$ & $16.3$ \\
        \textbf{SIM  } $\uparrow$ & $\textbf{92.7}$ & $91.7$  & $90.3$ & $90.1$ \\
        \bottomrule
    \end{tabular}
\end{table}

\subsection{Synthesis efficiency gain due to speechcode decoder}
The speechcode decoder is capable of streaming, i.e. generating speech incrementally. Combining this feature with the autoregressive SpeechGPT gives our system a first-byte latency as low as 100ms - only a few decoded speechcodes are sufficient to produce intelligible speech. This minimal latency contrasts with the diffusion-based decoder that requires the entire speech sequence (one or multiple sentences) to be generated in one go, where first-byte latency equals total generation time. Further, we observe that the speechcode decoder makes the overall system 3X times more compute efficient compared to the diffusion baseline. We run a benchmark where we generate 1000 utterances of around 20 seconds in duration on a NVIDIA® V100 GPU with batch size of 1. On average, it takes 69.1 seconds for 1-billion-parameter SpeechGPT with the diffusion decoder to complete synthesis, but only 17.8s for the same SpeechGPT with the speechcode decoder.

\section{Related work}
\label{sec:related_work}

\paragraph{Text-to-speech as language modeling.}
Casting TTS problem as next token prediction has gained popularity in recent years due to how easy it is to scale language models to large data and model sizes.

The first model using this approach was TortoiseTTS \cite{tortoise} released in 2022. It combines a GPT2-style speechcode language model with a diffusion decoder and an off-the-shelf vocoder, achieving remarkable few-shot capability. VALL-E \cite{wang2023neural} followed a similar approach: the model is scaled to 60k hours of speech data and uses a pre-trained audio encoder EnCodec for speechcode extraction \cite{DBLP:journals/corr/abs-2210-13438}. 
VALL-EX \cite{VALLEX} replicated VALL-E with 70k hours, with an additional cross-lingual speech-to-speech translation task in English and Chinese by using target language token as prompt. VioLa \cite{wang2023viola} extended VALL-EX with both text-to-text translation and speech-to-text recognition tasks. SpeechX \cite{wang2023speechx} extended VALL-E by adding noise suppression, target speaker extraction, clean and noisy speech editing, and speech removal tasks.

While VALL-EX, VioLA and SpeechX are versatile models reporting improved performance on word error rate and speaker similarity, no effects on core TTS aspects such as naturalness or prosody improvement were reported. VALL-E reported significant improvements over industry baselines, but we are unable to compare as the model is not publicly available. Here, we propose a multi-lingual and multi-speaker TTS leveraging 100K hours of data, with strong proven naturalness results in English and Spanish. Evaluations show improved speech naturalness compared to Tortoise-TTS. Qualitative linguistic analysis shows "emergent abilities" \cite{wei2022emergent} - BASE TTS can render complex prosody patterns, taking cues from texts without explicit labels for emotions.

\paragraph{Discrete speech representations.} %

In audio generative AI, the first acoustic encoders used in GPT-TTS are VQ-VAE \cite{vqvae}, SoundStream \cite{zeghidour2021soundstream} and EnCodec \cite{DBLP:journals/corr/abs-2210-13438}, which are deep learning audio compression techniques producing discrete acoustic token based on vector quantization. These acoustic encoders are shown to be superior to Mel-spectrograms \cite{DBLP:journals/corr/abs-2309-10922, yang2023towards}, but ignore semantic information and are unnecessarily complex due to lack of disentanglement \cite{huang2023repcodec}. AudioLM \cite{borsos2023audiolm} combines SoundStream tokens with semantic BERT tokens \cite{chung2021w2v}. SpeechTokenizer \cite{zhang2023speechtokenizer} and RepCodec \cite{huang2023repcodec} aim to capture semantics by adding self-supervised embedding prediction-based losses \cite{mohamed2022self}. BASE TTS builds acoustic tokens by leveraging semantic information from self-supervised embeddings and disentangling speaker information from the acoustic tokens. Then, it applies byte-pair encoding on these tokens to reduce memory requirement, allowing the model to train on longer sequences.

\paragraph{LTTS simplifies modeling.} 

Since Tacotron~2 \cite{tacotron2}, a successful TTS paradigm has been to separate speech creation into three sub-systems: (1) a Frontend responsible for text normalization, grapheme-to-phoneme conversion and  SSML tag handling e.g. to mark emphasis \cite{DBLP:journals/corr/abs-2307-07062}, (2) a Mel-spectrogram generator, inferring the amplitudes of the speech signal and responsible for the model expressivity, and (3) a dedicated Vocoder \cite{DBLP:conf/ssw/OordDZSVGKSK16, jiao2021universal}, generating the waveform by inferring the phase information. Expert knowledge can be induced to improve its performance such as using pre-computed features \cite{DBLP:conf/iclr/0006H0QZZL21}, coarse or fine-grained prosody information (\cite{Copycat}, and style control \cite{DBLP:conf/naacl/PrateekLBDLMRW19}. It has also proven successful with architectures such as Transformers \cite{li2019neural, DBLP:conf/iclr/0006H0QZZL21}, Flows \cite{kim2021conditional}, and Diffusion \cite{popov2021grad}.

These systems require a complex pipeline. Any error made during an earlier stage is propagated to the next. BASE TTS and other LTTS models are breaking these limitations. We simplify the data preparation by requiring only a large amount of audio, where texts can be obtained from a speech-to-text system, and require no phoneme extraction. GPT-based architectures \cite{radford2019language} have flourished by enabling versatile prompt-based task formulation, integration of expert feedback \cite{ouyang2022training}, and use as a foundational model with quick fine-tuning 
\cite{hu2022lora, fu2023effectiveness} e.g. on a specific task, a set of speakers or a new locale. BASE TTS shows that an end-to-end approach can achieve high expressivity on a few audio examples and in a multilingual setting, marking a high bar of quality in Spanish.

\paragraph{Contextual and emotional TTS.}
This usually requires separate, dedicated TTS sub-systems. Multiple approaches rely on predicting prosody representations using contextualized word embeddings \cite{copycat2, xin2023improving, wu2022self, DBLP:journals/corr/abs-2309-01576}. While prosody predictors of these models usually generate appropriate prosody, they often ignore strong text cues that would force a dramatic change e.g. in emotions or speech cues like shouting or whispering. Context-aware emotional TTS systems \cite{pan2021chapter, liu2023emotion, emotts, mukherjee2022text, schnell2021improving}
are even more limited. They usually favor a text-based emotion predictor coupled with an emotion controlable TTS system. These systems require high-quality recordings with a forced speaking style and annotated audio and text data, limiting their usefulness due to the sheer number of emotions expressible in human speech \cite{cambria2012hourglass}. BASE TTS benefits from being a language model which is both acoustic and semantically/syntactically aware. In this paper, we systemize an approach to produce and evaluate emergent contextual understanding of the text with a wide range of styles, without requiring supervised training or annotation.

\paragraph{BASE TTS has data efficiency built-in.}
Low-resource TTS has focused on reducing the required amount of high quality training data through e.g. voice conversion \cite{huybrechts2021low}, data generation \cite{DBLP:conf/icassp/LajszczakPKBBJN22}, modeling techniques \cite{DBLP:journals/corr/abs-2307-07062}, or recording script optimizations \cite{shamsi2020tts}. This stems from the early difficulty of ingesting a high volume of data from multiple speakers in potentially different styles or languages. A step up in that direction is zero-shot inference such as in Bark \footnote{\url{https://github.com/suno-ai/bark}} and \cite{casanova2022yourtts, wu2022adaspeech}, which aim to clone a voice with only a few seconds of recordings. Due to leveraging a substantially larger dataset, a bigger model, and a dedicated speaker encoder, BASE TTS and the recent LTTS models set a new standard in data efficiency.

\section{Conclusion}
We introduced BASE TTS, a GPT-style TTS system using novel SSL-based speechcodes as an intermediate representation and a speechcode decoder that offers an efficient, streamable alternative to diffusion. This is the largest model of its kind known to us, both in terms of parameters and training data. We demonstrated new state-of-the-art TTS results against baselines including Tortoise, Bark and YourTTS. We proposed a new way to measure textual understanding of TTS models, and showed that LTTS models built with 10K hours of data and 400 million parameters start to exhibit an advanced grasp of text that enables contextually appropriate prosody. From BASE TTS's strong performance on English and Spanish, we caught a first glimpse of a multilingual TTS approach that achieves high expressiveness, adapts to textual clues, is data efficient, uses only public domain data, and works for streaming TTS usecases such as voicing LLM outputs. Our approach points towards potential Scaling Laws \cite{scaling-law} of LTTS models, where an even larger amount of speech and other (text, image) data are needed to support multimodal objectives \cite{facebook-multimodal} and to break new grounds in TTS. 

Our approach still contains some limitations: a) BASE TTS occasionally produces hallucinations and cutoffs, where we produce either extra or incomplete audio than intended by the text. This is an inherent problem with the autoregressive LM approach, made worse by the misalignment between audio data and the ASR-generated text; b)
Selecting the right discrete representation for GPT-style TTS is crucial. More research is needed to establish how different properties of speechcodes translate into end-to-end system quality. We only report results for one speechcode configuration and leave more comprehensive study for future work.

\section{Ethical statements}

BASE TTS is a high-fidelity model capable of mimicking speaker characteristics with just a few seconds of reference audio, providing many opportunities to enhance user experiences and support under-resourced languages. An application of this model can be to create synthetic voices of people who have lost the ability to speak due to accidents or illnesses, subject to informed consent and rigorous data privacy reviews. However, due to the potential misuse of this capability, we have decided against open-sourcing this model as a precautionary measure. Further, we acknowledge the impact of speech data composition on the ability of the model to express the speech of linguistic, ethnic, dialectal, and gender minorities \cite{Kachel-gender, Gaither-black, dialect-minority}. We advocate for further research to a) quantify the impact of data composition; b) identify methods to combat potential biases and foster inclusivity in voice products.

\bibliographystyle{unsrtnat}
\bibliography{references}

\begin{appendices}

\section{Emergent abilities testset}
\label{appendix:emergent}

\subsection{Questions}
1. You went to the party, even though I explicitly told you not to?

2. There is another aircraft still in the air???

3. Now, seriously, you're saying I am the one to blame?

4. But she clearly doesn't want to?

5. To Hungary and back?

6. You're a copper?

7. What is Data Informed Diplomacy, with all its various manifestations?

8. What's really happening, and is there more than meets the eye?

9. How on earth is this Financial Report organized?

10. Where has Jason Abhisheki moved all the flowers to? 

11. What do we do in this situation, and what are the implications for Jordan's water supply?

12. But the Brexit question remains: After all the trials and tribulations, will the ministers find the answers in time?

13. Sorry, can you restate your name and address please? 

14. Here's the full story for today, would you like to learn more? 

15. Mr. Chairman, your highly anticipated interview with Channel 4 has turned into a catastrophe, hasn't it? 

16. Johnny boy, don't go around acting tough if you can't back it up, right? 

17. Are you in favor of the Latex usage policy or you're just sucking up to leadership?

18. Is it a bird, or is it a plane?

19. Madam, have you tried turning it off and on again? 

20. Were you the one with the hand-held camera or the one with a weird-looking android phone?

\subsection{Emotions}

1. Her hands shaking with excitement, Alice Monroe stuttered, "oh..I-I can't believe it! Is this really my acceptance letter to Harvard?" Marco cannot believe it either: "God damn it! How did you pull this off?"

2. A surge of anger flashed across the face of Matthew, as he bellowed, "You have crossed the line this time, and I won't stand for it any longer! Get out!"

3. Gazing at the panoramic view from a mountain in Iceland, Jeff Farahmand sighed deeply, whispering, "This... this is truly the face of the Divine. What more can I ask for?"

4. "You mustn't underestimate how profoundly I've missed your presence," Ito murmured, his eyes glistening with tears as he embraced his long lost sister. "You're finally back, but where do I find our lost years?"

5. "Oh my gosh! Are we really going to the Maldives? That’s unbelievable!" Jennie squealed, bouncing on her toes with obvious glee.

6. "I can confidently declare that this is the most exquisite chocolate cake my taste buds have ever had the pleasure to encounter!" Mo proclaimed, savoring every bite. He could not stop eating! 

7. A proud smile spread across his face as he softly said, "Son, your accomplishments fill my heart with such joy and pride." But then the smile suddenly ceased. Mike’s hearts were pounding like door knocks. His dad’s face now looks like that of the devil himself. 

8. Choking back sobs, Mahmoud whimpered, "I simply can't fathom a life without you by my side. Don't go!"

9. His voice trembled with palpable fear as he stuttered, "There's... there's a stranger at the window. Where the hell are you all waiting for?!"

10. Tears of joy trickled down her cheeks as she yelled, "Graduating as valedictorian... this is a dream come true!"

11. Jane's eyes wide with terror, she screamed, "The brakes aren't working! What do we do now? We're completely trapped!"

12. A profound sense of realization washed over Beal as he whispered, "You've been there for me all along, haven't you? I never truly appreciated you until now."

13. Beth collapsed into his arms, sobbing uncontrollably, "I failed them, I failed them all. They’re all dead! Nothing we can do will ever bring them back. How can I ever live with myself again? How?"

14. His face lit up with pure delight as he exclaimed, "We did it! We won the championship! I knew we could do it together!"

15. Overcome with guilt, Martin hung his head and muttered, "I'm so sorry. I never meant to hurt you like this. Can you ever forgive me?" It was obvious what the answer would be. 

16. The queen danced around the room, eyes twinkling with mischief, "Guess what? I got the lead role in the play! Can you believe it? Well, I can’t."

17. Staring into the distance, the firefighter said with a melancholic smile, "She used to sit right there, you know. I can still hear her laugh if I close my eyes." Outside the window, the rain was pouring down and gushing through every cracks. 

18. The detective’s voice, full of determination and fire, was heard loud and clear in the room, "No one will tell me what I can or cannot do. I'll prove them all wrong! Get me my gun. What are you all looking at me for?"

19. Overwhelmed with confusion and despair, David Darlan cried out, "What do you want from me? Why can't you just tell me what's wrong? Leave me alone!"

20. With a gentle touch and a loving smile, she reassured, "Don't worry, my love. We'll get through this together, just like we always have. I love you."

\subsection{Compound Nouns}

1. In the heart of Lagos, there is a public park with a serene duck pond. Nearby, the children's outdoor play area is full of joyful laughter. Nobody knows the darkness descending soon. 

2. At the family reunion, my grandfather, or father-in-law for some, told many tongue-in-cheek jokes. 

3. The physics teacher asked the students to build a new model solar system. Students were told to bring a tape measure and a pair of scissors, to cut the scale-model planet rings.

4. On this fateful day in 1987, the students boarded the little yellow school bus, chattering excitedly about their field trip to the zoo.

5. Hello, we are representatives from Northern Airlines. Please look out from the big second-floor window seat.

6. After years of work, Heisenberg finally published a ground-breaking cutting-edge research paper on quantum physics.

7. Recipe for a delicious breakfast sandwich: avocado, egg, and cheese on a bagel, cooked over a stovetop frying pan.

8. There is nothing more peaceful than a blue water fountain with a wooden greenhouse. Near there, Joseph installed a hard stone birdbath.

9. Prague, Czechia: Good morning, Harari! Here come the big shopping carts and last-minute video game shoppers.

10. My dog knocked over the tea table and all the books scattered across the second living room floor. 

11. The hiking trail up Yahu Mountain provides a spectacular view of the sunrise. Along the path, the wooden signposts with triple-checked trail maps and green distance markers guided us.

12. The fish clock tower was striking again, reminding us of that profound changing of the guard. 

13. Dean Graham sat on the packed wooden park bench, feeding the pigeons while enjoying the pleasant weather. 

14. The Beckhams decided to rent a charming stone-built quaint countryside holiday cottage. 

15. The construction of the new Newtown-council town hall has made huge trouble; rush-hour traffic jam has never been worse. 

16. Owen Farrell has taken England to the Rugby World Cup glory, with a razor-thin-margin victory against New Zealand in France.

17. Scientists at AWS teams are making last-minute pre-launch model preparations.

18. Bad weather in Northern Europe has caused a god-awful flight check-in time of 6 AM, when even the airport food court isn't open. 

19. Jake Park boasts a beautiful hand-built wooden bird feeder.

20. We visited a quaint bed-and-breakfast establishment, complete with lighthouse lamp room.

\subsection{Syntactic Complexity}

1. The complex houses married and single soldiers and their families.

2. Time flies like an arrow; fruit flies like a banana.

3. The rat the cat the dog chased killed ate the malt.

4. After the writer the editor the publisher hired fired quit, the company found itself in quite a bind.

5. The old man the boats on the shore were manned by had a long history of seafaring.

6. Anyone who feels that if so many more students whom we haven't actually admitted are sitting in on the course than ones we have that the room had to be changed, then probably auditors will have to be excluded, is likely to agree that the curriculum needs revision.

7. While John, who had been working late every night for a month on his novel, finally took a break to enjoy the fresh air, his neighbor, a painter who often found inspiration in the midnight moon, was just beginning her creative process.

8. In the old village with its winding roads, colorful marketplaces, a sense of history that permeates every brick, and a single traffic light, you'll find peace and simplicity.

9. The chef seasoning the fish tossed it gently.

10. As the sun dipped below the horizon, casting a golden glow over the ocean, Emily, who had spent her life dreaming of distant shores, stood on the deck of the ship, feeling a mixture of anticipation and nostalgia as her adventure began.

11. During the meeting, where Coke executives debated the future of the company, Thomas, a young intern who had discovered a solution, mustered the courage to speak, shifting the direction of the conversation, that preceded his intervention.

12. The movie that De Moya who was recently awarded the lifetime achievement award starred in 2022 was a box-office hit, despite the mixed reviews.

13. In the garden, where the flowers that the gardener who retired last year still bloomed, the children who play there every afternoon find peace and joy.

14. The scientist, Mateusz Gorka, who proposed the theory, which many experts in the field, including those who had initially been skeptical bordering on disbelieving, now support, was nominated for a prestigious award.

15. Although the meal that the chef, who had just returned from a culinary tour of Italy, prepared was delicious, the Greek guests barely noticed.

16. The book that the woman who the man who the child spoke to this morning was reading became a topic of conversation among the friends who had all read it.

17. Despite the fact that the road that led to the Five Villages, which was known for its scenic beauty, was narrow and winding, tourists flocked there throughout the year.

18. CNN journalists tracking the stories behind the officials who served during the tumultuous period when the protests rocked the nation to its core noticed significant inconsistencies in the official reports provided.

19. The musicians who performed the symphony that the composer, whose work had often been overlooked in his lifetime, wrote in his early years received a standing ovation.

20. Cars displayed in these showrooms with ENERGY-EFFICIENT AND GREEN decals prominently featured across the windshield aren't announcing environmentalism; they're virtue signaling.

\subsection{Foreign Words}

1. With an ample supply of joie de vivre, Mary danced through the streets of Nice, stopping only to enjoy a nice cafe with a warm croissant.

2. The modern exhibit was a mélange of styles, from German Expressionism to French Impressionism, capturing the Zeitgeist of the time.

3. As a gesture of camaraderie, the Spanish torero invited his rival, Leo the Monster, to a tapas bar, where they enjoyed jamón ibérico and the noche.

4. During Anthony’s wanderlust-filled travels, he discovered the gemütlich atmosphere of many Austrian villages.

5. CloudCorp’s CEO believes in gesamtkunstwerk, like integrating a symphony into a harmonious ensemble.

6. Mr. Henry, renowned for his mise en place, orchestrated a seven-course meal, each dish a pièce de résistance.

7. The fiesta, filled with música, dance, and the warmth of amigos, continued until dawn, embodying the true spirit of a Catalan celebration.

8. At the G20 Summit, leaders discussed rapprochement, trying to step away from the Schadenfreude of political rivalries.

9. After a tiring day, Sarah treated herself to a spa experience, enjoying the sauna and the jacuzzi, and relaxing with a glass of Riesling. 

10. Lasso's novella, rich in allegory and imbued with a sense of ennui, drew from his experiences living in a French château up near the border.

11. The master from Osaka, Japan, dedicated himself to crafting the perfect "nigiri," with "umami" flavors dancing on the palate.

12. Mikhail Gorbachev's Reforms: Perestroika and Glasnost Define a New Era.

13. Lakshmi's yoga practice, centered around the Sanskrit concept of "ahimsa," influenced her approach to life, mirroring the teachings of Mahatma Gandhi.

14. As they strolled through the Grand Bazaar in Istanbul, they were drawn to the beautiful "kilims," the best of Turkic craftsmanship.

15. Inspired by the ancient Chinese philosophy of "Feng Shui," Li rearranged her house to create a "qi" flow throughout.

16. Embracing the Japanese aesthetic of "wabi-sabi," Hokusai's masterpieces were on full display here.

17. During Rio de Janeiro's famous Carnaval do Brasil, the streets pulsated with the rhythms of "samba".

18. The novel's protagonist, guided by the ancient Greek concept of "arete," seeks excellence and virtue, a journey reminiscent of warrior-philosopher-kings.

19. As an aficionado of Scandinavian design, Ole Gunnarsson appreciated the principle of "hygge," evident in his Danish home.

20. These soldiers - they're supposed to practice with a sense of "bushido", the samurai code of honor, but they're behaving like the imperial beasts they are.

\subsection{Punctuations}

1. After a moment of silence, Elena Ivanova finally spoke..., —— her words barely audible over the cracking thunder of a torrential downpour.

2. What!?! You're telling me you've never seen a single episode of 'Game of Thrones' before????! (This was not heard by Prof. Johnson, Dr. Lewis, etc.)

3. "Can anyone hear me over there??? Please, we need help!!! NOW!!!!"

4. “The Power of \& and \% in the Digital Age.” won the first prize in this conference.

5. His latest invention (a device meant to assist in everyday chores (something he never seemed to run out of)), was nothing short of brilliant.

6. She read the label and was surprised to find --- that the "natural" ingredients were actually ..... heavily processed.

7. He relayed his conversation with the bartender, saying, "I told him, 'Your 'signature' cocktail is simply a Margarita with a fancy garnish.'"

8. The presently announced laws were announced in 35°N, 80°W. Specific provisions are to be found in §12 and §17.

9. Please ensure you replace [username] and [password] with your actual credentials before logging in, like jA8!fR3\$mQ1.

10. When Maria asked, 'What's happening tonight?' I replied, 'Well, John — who'll be there at 8:00 p.m. — said, "Let's meet at Sarah's place; bring games, snacks, etc., and don't be late!"'

11. "In the case of Johnson v. Smith, the court found that the defendant's actions — e.g., his failure to fulfill the terms of the contract (see sections 4.1, 4.2, and 4.3), etc. — amounted to a breach of trust.“

12. When asked for his thoughts, he simply replied, «I'll be gone in 5 minutes», and left.

13. I saw Gordon listing the ingredients as follows: <tomatoes>, <fresh basil> (or dried, if unavailable - but it's essential), <olive oil>, <garlic>; salt and pepper.

14. She received an odd text from her brother: 'Emergency @ home; call ASAP! Mom \& Dad are worried...\#familymatters.'

15. The sign at the park's entrance stated, 'Please adhere to the following rules: no littering; no pets (except service animals); no loud music after 9 p.m.'

16. “The Art of /Slash/ and $\backslash$backslash$\backslash$” was the best received talk on modern internet lingo.

17. Jeb's email was brief, but to the point: 'Meeting rescheduled for 3 p.m. tomorrow – apologies for any inconvenience. Regards, J.'.

18. The Dead Sea poems contained several annotations, some of which were quite puzzling: [Section unclear]; [Translation disputed]; [Original wording lost].

19. Her travel blog post was filled with enthusiastic descriptions: 'Best trip ever!!!'; 'Amazing people \& culture!'; 'Can't wait to go back...\#wanderlust.'

20. He shouted, 'Everyone, please gather 'round! Here's the plan: 1) Set-up at 9:15 a.m.; 2) Lunch at 12:00 p.m. (please RSVP!); 3) Playing — e.g., games, music, etc. — from 1:15 to 4:45; and 4) Clean-up at 5 p.m.‘

\subsection{Paralinguistics}

1. Principal Dickson began, addressing the Parkside assembly: "Ahem, I'd like to talk to you about something real serious."

2. "Aha! Now I understand," said Staff Sgt. Miller, piecing together the evidence. "The culprit left this behind. Phew."

3. "Ouch! That stings," Lilly cried, as her mother carefully applied the antiseptic. "Not beyond salvation, eh?" She dryly asked. 

4. "Shh, Lucy, sshhh, we mustn't wake your baby brother," Tom whispered, as they tiptoed past the nursery.

5. "Hmm, what do you think, is it too high or two low, um... Dr. Carter?" Haim asked, handing over the instrument. 

6. "Uh, well, Lisa," Tarek stuttered, nervously extending the ring he bought for god-knows how much, "mmm..will you marry me?"

7. "Yawn," Robert said, stretching out on the park bench, "this sunshine makes me sleepy."

8. "Oops! I did it again!" little Katie exclaimed, spilling her milk. 

9. "Whoa, can you believe this, Mike?" Susan said, staring at the intruder. "Wow, you're right. These men ain't meanin' well." 

10. James leaned back in his chair, wiped his forehead, and sighed, "Phew, haha, that was a tough meeting. Thanks for being there, Karen."

11. psst. psst. look right here. 

12. "Aha! I've found it, Professor Green," exclaimed Muzi Han, holding up the rare manuscript. "This could change our entire understanding of history."

13. "Ouch, be careful, Henry!" warned his sister, as he climbed the rickety ladder.

14. David whispered to Emily as the lights dimmed in the theater, "Shh, it's starting." 

15. "Hmm, I don't know about this, Jim," Mary said, looking at the folder paper. "It doesn't seem right."

16. "Uh, are you sure about this?" Tim asked nervously, looking at the steep slope before them. "Whoa, it's higher than I thought," he continued, his voice filled with trepidation. "Aha, but look at the view," Emily responded with excitement, "it's worth the climb!"

17. Ta-da! well? What do you think? This is the best right?

18. "Oops, sorry, Dad!" Jack apologized. "Ugh! you again". Dad was impatient. 

19. "Whoa, what a game, Alex!" Chris exclaimed. "I've never seen anything like that final play."

20. "Phew, we made it, Martha," Tom said, collapsing into the seat after the completion of the Manhattan Project.

\end{appendices}

\end{document}